# An Experimental Evaluation of Japanese Tokenizers for Sentiment-Based Text Classification

Andre Rusli   Makoto Shishido
Graduate School of Advanced Science and Technology
Tokyo Denki University
{20udc91@ms., shishido@mail.}dendai.ac.jp

## 1 Introduction

One popular utilization of machine learning text classification in the industry is sentiment-based classification. While there has been significant progress in sentiment analysis for high-resource languages, such as English, not much effort has been invested in analyzing Japanese due to its sparse nature and the dependency on large datasets required by deep learning [1]. In many languages, words are often considered as the basic unit of texts. Several works of research also show that when using n-gram language models, word n-grams are relatively better than character n-grams to convert texts into tokens when building a text classification model [2, 3].

Japanese texts pose different challenges for a machine learning algorithm to perform well. Sentences in Japanese contain no whitespace between words, so the common preprocessing phase to explicitly split words based on whitespaces could not be easily conducted. In addition, the combination of characters in a sentence could vary and may be ambiguous as they could have different meanings depending on the combinations. Many works of research have developed morphological analysis tools for Japanese language. Some of the popular tools for morphological analysis tools for Japanese are including MeCab [4, 5], Sudachi [6], and SentencePiece [7].

In this article, we focus on utilizing one of the many features provided by the previously mentioned tools which is the tokenizer for segmenting Japanese texts by words or subwords. Furthermore, this research then implements the tokenizer as a preprocessing step towards building supervised sentiment-based text classification models with Term Frequency–Inverse Document Frequency (TF-IDF) vectorization. We use Multinomial Naïve Bayes and Logistic Regression to build the classification model and to provide comparison between two popular traditional machine learning algorithms which are often used as baseline classifiers. Moreover, relatively low model complexity and higher level of interpretability [8], when compared to advanced deep learning algorithms, are also preferable.

## 2 Related Works

MeCab, Sudachi, and SentencePiece as morphological analysis tools have been used in many works of research in natural language processing. Although it is not exactly a newly proposed tool (the last update on its GitHub repository is version 0.996 in February 2013 [4]), MeCab is still used extensively as a word segmentation tool for preprocessing Japanese text in recent years. One example is the work by Zhang and LeCun [9], in which MeCab is used to segment texts in Japanese and Korean (with additional model in Korean language). Sudachi is another tool that mainly emphasizes the focus on continuous maintenance and feature richness, as it aims to support business use. It is also currently used by spaCy [10], a well-known library in NLP, as a tool for their pretrained statistical model for Japanese.

SentencePiece, on the other hand, is a tool for subword segmentation that uses a different approach. The authors describe SentencePiece as a language-independent subword tokenizer and detokenizer designed for Neural-based text processing [7]. While using a different approach with focus on providing a method to support language-independent multilingual text processing, its performance is shown to be effective for various tasks, such as English-Japanese neural machine translation [7],

Japanese news text classification [11], and sentiment analysis in Japanese [1]. However, even though many articles have shown the ability of each tools to be utilized for various tasks in natural language processing, works that provide a hand-in-hand comparison between the features provided by the tools in the same environment and configurations are still hard to find.

Regarding text classification tasks for Japanese texts, many recent works experiment with complex models using various deep learning approaches such as BERT [1, 11], Bidirectional LSTM-RNN [12], and Quasi-RNN and Transformer model [13], most are competing to achieve state-of-the-art performance. In our early work presented in this article, we aim to experiment with and report text classification results on baseline traditional algorithms with less computational cost and more interpretability.

Two classifiers used in our work are the Multinomial Naïve Bayes and Logistic Regression. Logistic Regression, which used to be the default choice for text classification, can be found used as a baseline linear model for Japanese text classification such as in [1, 9, 14]. Many researchers are also still exploring the potential and possibility of enhancing Multinomial Naïve Bayes' ability as a classification method, such as in [15]. By using these two methods, we emphasize more on the tokenization tools, rather than the classification algorithms. In addition, to also provide reports on the performance of traditional machine learning algorithms which with its limitations could still perform well. The goals of our current study are summarized as follows:

- Experiment with tokenizers provided by MeCab, Sudachi, and SentencePiece, then use the tokenization results to build models for binary sentiment-based text classification using TF-IDF with Multinomial Naïve Bayes and Logistic Regression.
- Compare and report the performance results in terms of time and error percentages.

## 3 Experimental Setup

### 3.1 Dataset

Our study is based on the Japanese Rakuten product review binary sentiment dataset provided in [9]. The datasets (both train and test set), which are available as CSV files, consist of three columns supposedly the binary sentiment label, review title, and review text, with examples such as the followings:

- Label: 1 (negative)
  Review title: 臭い
  Review text: 余りにも、匂いがきつく安物みたいです。\\n 安いから仕方ないかな？
- Label: 2 (positive)
  Review title: 早いし安い
  Review text: 毎回利用しています。納品が早いし何よりお安く大変便利です。また利用します。

We use the binary sentiment label and review text and we randomly sampled 10% of the provided training and testing data and only used the sampled 10% of the total data in our experiments, that is 340,000 reviews for training and 40,000 reviews for testing as we aim to experiment with many configurations in a limited setting.

### 3.2 System Specifications

The experiments are conducted by using Python 3.7 and Jupyter Notebook, running on an Ubuntu virtual machine provided by Google Compute Engine, using a machine type c2-standard-4 with 4 vCPUs and 16 GB memory. However, it is also possible to run all of the experiments on Google Colaboratory. For the tokenization tools, we use the available MeCab (mecab-python3 1.0.3), Sudachi (SudachiPy 0.5.1), and SentencePiece (sentencepiece 0.1.94) packages available via the Python Package Index (PyPI) [16]. As for the TF-IDF vectorizer, Multinomial Naïve Bayes classifier, and Logistic Regression classifier, we used packages provided by Scikit-learn [17, 18].

### 3.3 Methods

After preparing our environment specifications as previously described, we could then proceed to conduct our experiments. The overall flow of our experiments can be seen in Fig. 1. After the original Rakuten binary dataset is downloaded, firstly, we randomly sampled ten percent of the provided train and test data, resulting in a total of 340,000 train data and 40,000 test data.

We then proceed to train a SentencePiece (SP) model based on the sampled train data, setting the vocab_size parameter as 32,000. We initially chose 32,000 based on the result shown in experiments in a previous work by

Yohei Kikuta [11].

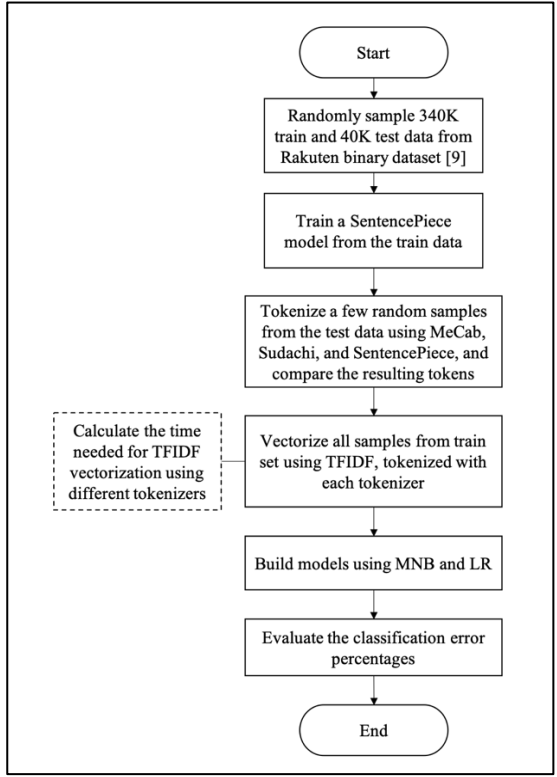

Fig. 1. Flow of the experiments conducted in our study

Unlike SentencePiece, Sudachi has its own dictionary with different sizes and MeCab can be integrated with various existing dictionaries in order to perform tokenization, so there is no need for more pre-training. We use Sudachi's core dictionary for surface-form tokenization using Sudachi, and we use the unidic-lite dictionary for tokenization with MeCab. Furthermore, before building our classification model, we experimented with few randomly selected reviews to get a glimpse of the tokenization results by the three different tokenizers.

After that, using each tokenizer, we vectorize all the reviews in the training set to create matrices of TF-IDF values which will then be used to build our classification model. In this process, we also calculated the time spent by each tokenizer to process various number of train data into their TF-IDF values. We then use the vectorized data to build two classification models for each tokenizer, one using the Multinomial Naïve Bayes (MNB) classifier, and another one using the Logistic Regression (LR) classifier. Finally, we test our models to classify reviews in the test set and evaluate the error percentages.

## 4 Results and Analysis

### 4.1 Tokenization Result

We selected several review texts from the train and test set, then try to segment the words or subwords in order to experiment with each tokenizer. Vocab_size=32,000 is used for the SentencePiece model. Some parts of review texts along with their tokenization results are as follows.

1. Review text: "自転車通勤用に購入。サックスを選びましたが、…"
   a. MeCab: ['自転', '車', '通勤', '用', 'に', '購入', '。', 'サックス', 'を', '選び', 'まし', 'た', 'が', …]
   b. Sudachi: ['自転車', '通勤用', 'に', '購入', '。', 'サックス', 'を', '選び', 'まし', 'た', 'が', …]
   c. SentencePiece: ['▁', '自転車通勤', '用に購入', '。', 'サックス', 'を選びましたが', …]
2. Review text: "かわいいです(*^。^*)\n パソコン…"
   a. MeCab: ['かわいい', 'です', '(', '*^。^*)\\', 'n', 'パソコン', …]
   b. Sudachi: ['かわいい', 'です', '(*^。^*)', '\\', 'n', 'パソコン', …]
   c. SentencePiece: ['▁', 'かわいいです', '(*^。^*)', '\\', 'n', 'パソコン', …]

The examples provided above are randomly selected and might not be fully representative to showcase the full capabilities of each tokenizer, however, some general similarities and differences can be observed. Compared to MeCab and Sudachi, tokens generated by SentencePiece are not based on any formal dictionary, so there are words or subwords that does not match formal Japanese dictionary, for example the token "便利ですね" ("benridesune") which is usually segmented into three words ("benri", "desu", and "ne") but treated as one token. Moreover, although tokens generated by MeCab and Sudachi are generally similar, there are some characters that are treated different depending on the context they are in. For example, the words "自転車通勤用" and the combination of characters comprising the emoji "(*^。^*)", are segmented differently in MeCab and Sudachi. Sudachi could divide the characters into ["自転車"(bicycle), "通勤用"(for commuting)] and treat the whole emoji as one union, while MeCab further separate the combination into ["自転", "車", "通勤", "用"] and

[" (", "*^｡ ^*)\\"].

## 4.2 TF-IDF Vectorization

We then proceed to use each tokenizer to vectorize the all texts in our train data (340,000 review texts) using TF-IDF and calculate the time elapsed. As can be seen in Table 1, SentencePiece is the fastest while MeCab is only slightly slower, and Sudachi needs the longest time. This might be caused by Sudachi's focus in providing high quality segmentation based on its continuously updated rules and dictionary that enable it to divide words as can be seen in the previous paragraph.

Table 1. Elapsed time to vectorize TF-IDF values

| Tokenizer | Elapsed Time (seconds) |
|---|---|
| MeCab | 34,65 |
| Sudachi | 1533,58 |
| SentencePiece | 25,84 |

## 4.3 Text Classification Models

After observing the tokens generated by each tokenizer and calculated the time elapsed to vectorize TF-IDF values, we proceed to use the TF-IDF values generated by each tokenizer and train two models using Logistic Regression (LR) and Multinomial Naïve Bayes (MNB). We could observe the effect of each tokenizer, with varying tokenization approaches and time elapsed to generate the TF-IDF values, on the classification performances in Table 2.

Table 2. Classification error percentages

| Tokenizer | Classifier | Error-Train (340,000) | Error-Test (40,000) |
|---|---|---|---|
| Mecab | LR | 8.52 | 9.63 |
| Sudachi | LR | 8.5 | 9.59 |
| SP | LR | 6.54 | 8.02 |
| Mecab | MNB | 11.24 | 12.52 |
| Sudachi | MNB | 11.04 | 12.42 |
| SP | MNB | 8.28 | 8.9 |

In our experiment using linear model such as Logistic Regression, and also Multinomial Naïve Bayes, with TF-IDF vectorizer, the combination of SentencePiece with Logistic Regression outperforms the others with 6.54 error percentage on the training set and 8.02 error percentage on the testing set. We can see that even though the segmentation results by SentencePiece are quite different with the other tokenizers' results, it seems to work well on a linear classifier to solve binary classification problem. Another finding is that in our task, even though Sudachi, with its 'surface form' tokenization, could perform relatively better word segmentation, the resulting classification model could perform only slightly better than MeCab, despite taking the longest elapsed time in the vectorization phase.

Furthermore, using the best performing model in Table 2 (Logistic Regression with SentencePiece), we then perform a hyperparameter tuning process for Logistic Regression using grid search and repeated stratified k-fold cross validator from Scikit-learn. Table 3 shows the error percentages of our final model (train error: 6.54, test error: 8.02) using the following hyperparameters on the Logistic Regression classifier: C=10, penalty='l2', solver='lbfgs'.

Table 3. Error percentages after hyperparameter tuning

| Tokenizer | Classifier | Error-Train | Error-Test |
|---|---|---|---|
| SP | LR | 5.56 | 7.78 |

## 5 Conclusion and Future Work

Our experiments showed the results of an experimental evaluation of three popular tokenization tools, MeCab, Sudachi, and SentencePiece, for processing Japanese texts. The resulting tokens are then used to train text classification models using TF-IDF with Logistic Regression and Multinomial Naïve Bayes. We found that the generated tokens from Sudachi are more likely to match dictionary results and common words understood by human, however, MeCab and SentencePiece are significantly faster. Moreover, even though tokens generated by SentencePiece are limited to its training data and might not match common dictionary results, they perform better for our dataset, which is a binary sentiment-based text classification task. Finally, the combination of SentencePiece, TF-IDF, and Logistic Regression achieved the best performance with 5.56 training error percentage and 7.78 testing error percentage.

This article reported the result of an ongoing research work. Some future research steps include experimenting with various n-gram configurations and other hyperparameters, using multi-class and bigger datasets, training multi-lingual models, and experimenting with various shallow and deep learning approaches.